\documentclass{article}
\usepackage{spconf,amsmath,graphicx}
\usepackage{booktabs}

\usepackage{color,balance, cite, bm} 

\def \eg {\emph{e.g.}}
\def \ie {\emph{i.e.}}

\title{Dynamic Texture Recognition Using PDV Hashing and Dictionary Learning on Multi-Scale Volume Local Binary Pattern}
\name{Ruxin Ding, Jianfeng Ren, Heng Yu, Jiawei Li\thanks{This work was supported in part by the National Natural Science Foundation of China under Grant 72071116, and in part by the Ningbo Municipal Bureau Science and Technology under Grants 2019B10026.}}
\address{School of Computer Science\\ 
		University of Nottingham Ningbo China\\
		199 Taikang East Road, Ningbo, 315100 China}
\begin{document}
%
\maketitle

\begin{abstract}
\small{Spatial-temporal local binary pattern (STLBP) has been widely used in dynamic texture recognition. STLBP often encounters the high-dimension problem as its dimension increases exponentially, so that STLBP could only utilize a small neighborhood. To tackle this problem, we propose a method for dynamic texture recognition using PDV hashing and dictionary learning on multi-scale volume local binary pattern (PHD-MVLBP). Instead of forming very high-dimensional LBP-histogram features, it first uses hash functions to map the pixel difference vectors (PDVs) to binary vectors, then forms a dictionary using the derived binary vector, and encodes them using the derived dictionary. In such a way, the PDVs are mapped to feature vectors of the size of the dictionary, instead of LBP histograms of very high dimension. Such an encoding scheme could extract the discriminant information from videos in a much larger neighborhood effectively. The experimental results on two widely-used dynamic textures datasets, DynTex++ and UCLA, show the superior performance of the proposed approach over the state-of-the-art methods.}
\end{abstract}

\section{Introduction}
\label{sec:intro}

\begin{figure*}[htbp] 
	\centering
	\includegraphics[width=0.9\textwidth]{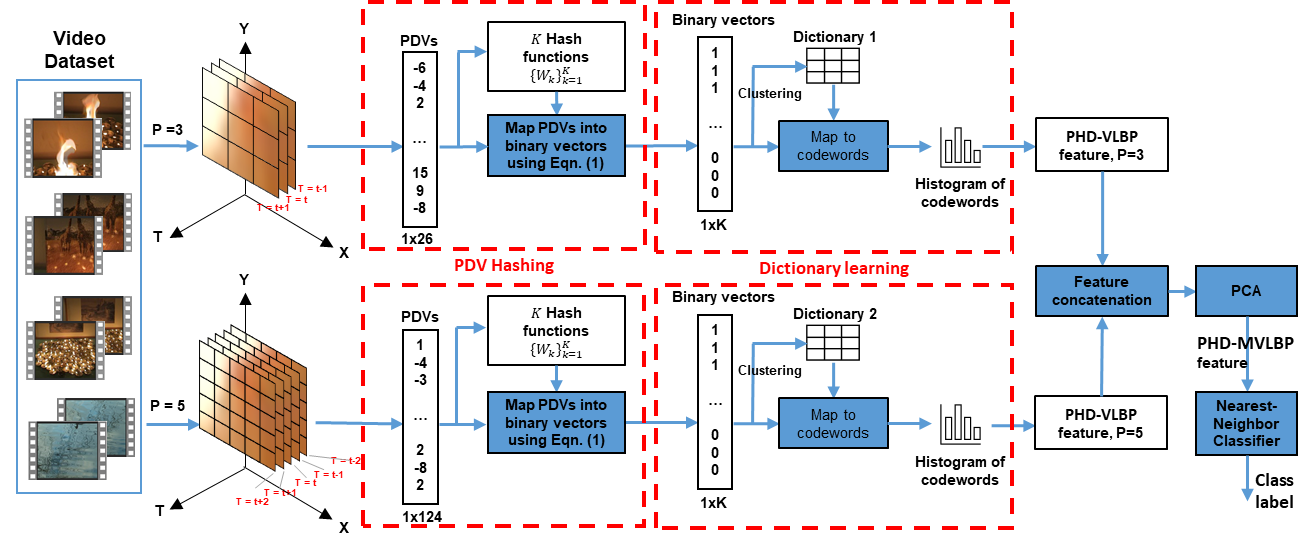}
	\caption{Overview of the proposed method. Pixel difference vectors (PDVs) are first generated by comparing neighbors with the center pixel. The PDVs are then mapped to binary vectors using hash functions, where the binary vectors and hash functions are jointly optimized. The generated binary vectors are then clustered to form a codebook, and each vector is encoded using the derived dictionary. Finally, the nearest-neighbor classifier with cosine distance is used to classify dynamic textures.}
	\label{fig:block_diagram} 
\end{figure*}

Dynamic textures (DTs) refer to sequences of the image that consists of repeated patterns related to time and space. 
DT has been widely used in various applications such as video retrieval \cite{RelevanceFeedback}, fire detection \cite{5208233}, and micro-expression analysis \cite{7851001}. Compared to static textures, DT classification poses increased challenges because their appearance, organization and motion information are not static but time-varying \cite{8600380}. Consequently, a descriptive feature representation is a key to the success of dynamic texture recognition.

Many methods have been developed for dynamic texture recognition, \eg, geometric properties computation \cite{QUAN201785}, local spatio-temporal filtering \cite{Ramirez2015}, dictionary learning~\cite{7780409}, deep convolutional neural network~\cite{QI20161230, shervin2017dynamic}, and various spatial-temporal local binary patterns~\cite{4160945, MBSIFTOP, LPQTOP,ren2013dynamic, ren2015chi, ren2015learning}. Among these, STLBP is most widely used in DT recognition. Zhao et al. developed Volume Local Binary Pattern (VLBP) \cite{4160945}, which combines the motion and appearance features together, instead of analyzing each frame individually. However, the feature dimension of VLBP increases exponentially with the number of neighbors, which prevents VLBP from utilizing the information in a large neighborhood. To reduce the feature dimension, LBP-TOP \cite{4160945} was developed to extract LBP features in three orthogonal planes. Inspired by LBP-TOP, various improved STLBPs have been developed, \eg, MBSIF-TOP~\cite{MBSIFTOP}, LPQ-TOP~\cite{LPQTOP}, and ASF-TOP~\cite{Hong2018}.

LBP-TOP and its variants partially address the high-dimension problem of VLBP, but the resulting dimension may be still high, \eg, the dimension of LBP-TOP is $3\times2^{P^2-1}$ for a VLBP neighborhood of size $P\times P \times P$. In addition, as LBP features are extracted independently from the three orthogonal planes, the correlation information among the three planes is lost, which leads to possible performance degradation. 

To address these challenges, we propose a method using PDV hashing and dictionary learning on multi-scale VLBP (PHD-MVLBP). Firstly, the pixel difference vector (PDV) of neighbors w.r.t. the center pixel is mapped to a binary vector using hash functions. Both the hash functions and binary vectors are jointly optimized so that the resulting binary vectors are evenly distributed. Human knowledge on image/video such as the uniformity of LBP codes~\cite{ren2013noise} is embedded in the optimization function. Then, clustering is performed on the binary codes to construct a dictionary, and a histogram feature for each video clip is derived using the learned codebook. Multi-scale histogram features are extracted for neighborhoods of different sizes. In such a way, the learned feature representations aggregate the discriminant information using all neighbors in the LBP neighborhoods, at different scales, from all videos in the dataset.

The proposed PHD-MVLBP is compared with the state-of-the-art methods on DynTex++ and UCLA datasets for dynamic texture recognition. It consistently outperforms all the compared methods.

\section{Proposed method}
\label{sec:format}

\subsection{Problem Analysis of VLBP and LBP-TOP}
Local Binary Pattern is a popular descriptor to represent the local texture of an image~\cite{1017623, ren2013noise, ren2015quantized, ren2017sound, 7936534}. To extend it to the spatial-temporal textures, volume local binary pattern (VLBP), was introduced by Zhao and Pietikainen \cite{4160945}. The captured LBP features are extended from the two-dimensional plane to three-dimensional space, to capture the dynamic texture information. VLBP works well when analyzing video data since it combines appearance and motion information \cite{2survey}. But VLBP suffers from the problem of high dimensionality. Given a volume neighborhood of $P\times P \times P$, its feature dimension is as high as $2^{P^3-1}$. To tackle this problem, LBP-TOP~\cite{4160945} was developed, but its dimension is still as high as $3\times2^{P^2-1}$. Several variants such as MBSIF-TOP~\cite{MBSIFTOP}, LPQ-TOP~\cite{LPQTOP}, and ASF-TOP~\cite{Hong2018} may have similar high-dimension issues. As a result, $P$ is often limited to 3 in practice, which limits the power of LBP feature descriptors. Furthermore, as LBP features are extracted in three orthogonal planes independently, the correlation information among these three planes is lost, which may lead to a degradation in classification performance, as evidenced later in experiments.

\subsection{Overview of Proposed PHD-MVLBP}
The proposed PHD-MVLBP aims to tackle the challenges of VLBP and LBP-TOP, \ie, effectively extracting the discriminant information from a volume of neighborhoods, without exponentially increasing the feature dimension and losing the discriminant information by splitting into three orthogonal planes. Towards this end, instead of constructing the LBP histogram features of high dimension, the whole pipeline is redesigned, as shown in Fig.~\ref{fig:block_diagram}. Firstly, pixel difference vectors (PDV) are extracted from local volume neighborhood by comparing neighbors with the center pixel. Instead of directly thresholding the PDVs into binary vectors in traditional LBP~\cite{4160945}, PDVs are mapped to binary vectors using hash functions, and binary vectors and hash functions are jointly optimized. The optimization functions are carefully designed so that human knowledge on LBPs are well incorporated into the formulation. Secondly, the derived binary vectors are clustered to form the dictionary and each vector is encoded using the derived dictionary. Thirdly, multi-scale features are extracted using the neighborhood of different sizes. Finally, a simple nearest-neighbor classifier with cosine distance is used to evaluate different LBP feature descriptors.

\subsection{Mapping PDVs into Binary Vectors Using Hash Functions}

Firstly, PDVs are extracted by comparing neighbors with the center pixel within a local neighborhood. Formally, give the neighborhood of 
size $P\times P \times P$, the center pixel is $I_c$ and the neighboring pixels are $I_1, I_2, \dots, I_{P^3-1}$, respectively, the pixel difference vector $\bm{x} = [I_1, I_2, \dots, I_{P^3-1}] - I_c \in \mathcal{R}^{P^3-1}$. Iterating this process for all neighborhoods in all videos could generate the PDVs $\bm{X} = [\bm{x}_1, \bm{x}_2,\dots, \bm{x}_N]$, where $\bm{x}_n$ is the $n$-th pixel difference vector and $N$ is the number of PDVs in the dataset. PDVs are capable of encoding significant micro-patterns, such as edges and lines. 

Secondly, we aim to map the PDVs into binary vectors using hash functions. Inspired by Duan et al.~\cite{7936534}, $K$ hash functions are used to map each $\bm{x}_n$ into a binary vector $\bm{b}_n = [b_{1n},…, b_{kn}]^T \in \{0,1\}^{K \times 1}$. The $k$-th binary code $b_{kn}$ of $\bm{x}_n$ can be computed as,
\begin{equation}
b_{kn} = 0.5 \times (\emph{sgn}(\bm{w}_{k}^{T}\bm{x}_n)+1),
\end{equation}
where $\bm{w}_k \in \mathcal{R}^{P^3-1}$ is the projection vector for the $k$-th function, $\emph{sgn}(v)$ equals to $1$ if $v \geq 0$ and $-1$ otherwise. 

Based on the concept of uniformity in LBP~\cite{ren2013noise}, the number of transitions between code $0$ and $1$ should be minimized, and hence the adjacent bits should be as equal as possible in the generated binary vector. However, such restraint could compel the learned binary code to all-zeros or all-ones, reducing the discriminant power of binary codes. To address this issue, we limit the sum of bitwise $0$ or $1$ switches in each binary code. Even if small alternatives occur in the original code $\bm{X}$, the learned binary vector could still be stable with this restriction. The objective function of feature representation is generated as follow, 
\begin{align}
\label{eqn:objective} 
    min_{\bm{w}_k} \emph{J} &= \emph{J}_1 + \lambda_1\emph{J}_2 + \lambda_2\emph{J}_3 + \lambda_3\emph{J}_4 \\ \nonumber
    &   = \sum_{n=1}^{N} {\Vert \sum_{k=1}^{K-1} {\Vert b_{kn} - b_{(k+1)n} \Vert}^2 - 1 \Vert^2}  \\ \nonumber    
    &   + \lambda_1 \sum_{n=1}^{N} \sum_{k=1}^{K} {\Vert (b_{kn} - 0.5) - \bm{w}_{k}^{T}\bm{x}_n \Vert^2} \\ \nonumber
    &   + \lambda_2 \sum_{k=1}^{K} {\Vert \sum_{n=1}^{N} (b_{kn} - 0.5) \Vert^2} \\ \nonumber
    &   - \lambda_3 \sum_{n=1}^{N} \sum_{k=1}^{K} {\Vert b_{kn} - \mu_k \Vert^2}
\end{align}
where $\mu_k$ is the mean of the $k$-th bit of all $N$ PDVs, and $\lambda_1$, $\lambda_2$ and $\lambda_3$ are three parameters to balance the weight of different terms. In our experiments, we set $\lambda_1 = 1000$, $\lambda_2 = 100$ and $\lambda_3 = 1000000$ empirically. The minimization of the first term, $\emph{J}_1$, attempts to make the adjacent bits of the learned binary codes as equal as possible and prevents all zeros or ones from showing up in the codes. By doing so, the learned codes could be more robust to noise. $\emph{J}_2$ is designed to minimize the loss of energy during the projection process via reducing the quantization loss. In order to make more information to be present, the goal for $\emph{J}_3$ is to evenly distribute each feature bit in the learned binary code. Furthermore, $\emph{J}_4$ could maximize the variance of binary codes to improve the independency of projection vectors, so that the generated binary vectors could carry as diversified information as possible. To find the most appropriate $\bm{W} = [\bm{w}_1,\bm{w}_2, \dots, \bm{w}_K]$, a gradient descent method is applied with the curvilinear search algorithm. More specifically, the objective function in Eqn.~(\ref{eqn:objective}) is optimized iteratively, by fixing $b_{kn}$ while optimizing $\bm{w}_k$ and by fixing $\bm{w}_k$ while optimizing $b_{kn}$. In the beginning, $\{\bm{w}_k\}_{k=1}^K$ is initialized as the top $K$ eigenvectors of $\bm{X}\bm{X}^T$. The algorithm terminates after a fixed number of iterations.

\subsection{Dictionary Learning for Binary Vectors}
After deriving the binary vectors, they are clustered to form the dictionary of $D$ codewords. Then, each vector is mapped to the nearest codeword. The histogram of codewords is used as the feature representation. In such a way, instead of being mapped to high-dimensional LBP-histogram features, PDVs are mapped to binary vectors using hash functions first and then mapped to histogram of codewords using the derived dictionary. Finally, principal component analysis (PCA) is applied to further compress the features and reduce the feature dimension. The derived features are less sensitive to illumination variations and local alterations with stronger discriminative power.

\subsection{Multi-scale Feature Extraction}
The proposed PHD-VLBP provides an effective scheme to encode the PDVs of a large neighborhood. To make full use of the discriminant information embedded in the dynamic texture videos, we propose a multi-scale scheme. We extract features using neighborhoods of different $P$. The PDVs of different scales are extracted and mapped to binary vectors of each scale correspondingly. Then, a dictionary is learned for each scale, and features extracted at different scales are concatenated. PCA is applied to the combined features to derive the final multi-scale features.  

\section{Experimental Results}
\label{sec:typestyle}

The proposed method is compared with the state-of-the-art methods on DynTex++ and UCLA datasets for dynamic texture recognition. The evaluation results are given in the following subsections.

\subsection{DynTex++ dataset}
The DynTex++ dataset \cite{ghanem2010maximum} consists of a set of DT videos, sampling from videos of the original DynTex dataset \cite{dyntex}. It categorizes the dynamic textures into 36 classes, where each contains 100 videos. The size of the video texture is $50\times50\times50$. For a fair comparison to existing methods, the experimental settings in \cite{6920055} are followed, where half of the dataset is randomly selected as the training set and the rest for testing. The experiment is repeated 5 times and the average result is reported. VLBP~\cite{4160945} and LBP-TOP~\cite{4160945} are closely related to the proposed method and chosen as the baseline methods. MBSIF-TOP~\cite{6920055} achieves previous best results and CVLBC~\cite{8030131} was published in a reputed journal recently, and hence they are chosen for comparison as well.

Two scales, $P=3$ and $P=5$, are used for our method, corresponding to a neighborhood of size $3 \times 3 \times 3$ and $5\times 5 \times 5 $, respectively. We report the results for these two scales and PHD-MVLBP. The dictionary size and PCA dimension are empirically set to 1500 and 500. The comparison results are summarized in Table~\ref{dyntex}. Compared to VLBP~\cite{4160945} and LBP-TOP \cite{4160945}, the performance gain of the proposed PHD-MVLBP is 10.42\% and 4.57\%, respectively, which demonstrates the effectiveness of the proposed method in extracting the discriminant information in videos. Compared to the previous best method, MBSIF-TOP \cite{6920055}, the performance gain is 0.6\%. The proposed PHD-MVLBP also outperforms two single-scale PHD-VLBP methods.
\begin{table}[]
\centering
\caption{Comparison results between the proposed method and other approaches on the DynTex++ dataset.}
\resizebox{.7\columnwidth}{!}{
\begin{tabular}{cc}
\toprule
\textbf{Method}           & \textbf{Accuracy}  \\ \midrule
Distance Learning \cite{ghanem2010maximum} & 63.70\%                         \\
DFA \cite{10.1109}               & 89.90\%                              \\
VLBP \cite{4160945}                             & 87.35\%                              \\
LBP-TOP \cite{4160945}           & 93.20\%                              \\
CVLBC \cite{8030131}             & 91.31\%                               \\
MBSIF-TOP \cite{6920055}         & 97.17\%                            \\ \midrule
Proposed PHD-VLBP, $P=3$          & \textbf{97.51\%}                    \\ 
Proposed PHD-VLBP, $P=5$          & 97.10\%                   \\ 
Proposed PHD-MVLBP          & \textbf{97.77\%}                    \\ \bottomrule
\end{tabular}
}
\label{dyntex}
\end{table}

\subsection{UCLA Dataset}
The UCLA dataset \cite{4270021} is a popular dataset for dynamic texture recognition. It consists of 200 DT sequences in total, including 50 scenes, with 4 sequences for each scene. The example dynamic textures are waterfalls, plants, swaying flowers, fire, boiling water, and fountains. For each sequence, there are 75 frames with $160 \times 110$ pixels. In our experiment, the dataset is cropped to contain the most motion among all videos with the size of $75 \times 48 \times 48$. Because of the insufficient data in the UCLA dataset to train a robust projection matrix and dictionary, these two trained on the DynTex++ dataset, a much larger dataset, are used on the UCLA dataset.

\subsubsection{50-Class Breakdown}
4-fold cross-validation is used, same as in \cite{6920055, MBSIFTOP, ghanem2010maximum}. As shown in Table~\ref{50class}, many existing methods achieve almost perfect classification results on the UCLA-50 dataset. The proposed methods, PHD-VLBP for $P=3$ and $P=5$, and PHD-MVLBP, all achieve the perfect classification accuracy and outperform all the compared methods.
\begin{table}[]
\centering
\caption{Comparisons between the proposed methods and other approaches on the UCLA dataset with 50-class setting.}
\resizebox{.7\columnwidth}{!}{
\begin{tabular}{cc}
\toprule
\textbf{Method}           & \textbf{Accuracy}  \\ \midrule
Distance Learning \cite{ghanem2010maximum} & 99.0\%                             \\
KDT-MD \cite{4270021}            & 97.5\%                              \\
CVLBC \cite{8030131}             & 99.5\%                              \\ 
MBSIF-TOP \cite{6920055}         & 99.5\%                             \\ \midrule
Proposed PHD-VLBP, $P=3$          & \textbf{100.0\%}                     \\ 
Proposed PHD-VLBP, $P=5$         & \textbf{100.0\%}                     \\ 
Proposed PHD-MVLBP          & \textbf{100.0\%}                     \\ \bottomrule
\end{tabular}
}
\label{50class}
\end{table}

\subsubsection{9-Class Breakdown}
In the UCLA database, each scene is often captured several times. Thus, the data in the UCLA dataset can be categorized into nine classes: 8 videos for boiling water, 8 for fire, 12 for flowers, 20 for fountains, 108 for plants, 12 for sea, 4 for smoke, 12 for water, and 16 for waterfall. Similarly, as in~\cite{6920055}, half of the data are randomly chosen as the training, and the rest are used for testing. The experiment is repeated 20 times. To make full use of the information embedded in videos, a video is divided into 5 non-overlapping sub-videos  with 15 frames in each sub-video. Majority vote is used to combine the classification results of sub-videos. 
The proposed PHD-VLBP for $P=3$ achieves an accuracy of 98.65\%, which significantly outperforms most of the compared approaches such as VLBP~\cite{4160945} and LBP-TOP~\cite{4160945},  and it is comparable to the multi-scale feature descriptor, MBSIF-TOP \cite{6920055}. The proposed PHD-MVLBP slightly outperforms the previous almost best-performed method, MBSIF-TOP \cite{6920055}, by 0.15\%. All these results demonstrate the superior performance of the proposed feature descriptors over the state-of-the-art methods for DT recognition. 
\begin{table}[]
\centering
\caption{Comparisons between the proposed method and other approaches on the UCLA dataset with 9-class setting.}
\resizebox{.9\columnwidth}{!}{
\begin{tabular}{cc}
\toprule
\textbf{Method}           & \textbf{Recognition accuracy rate}  \\ \midrule
Distance Learning \cite{ghanem2010maximum} & 95.60\%                               \\
VLBP \cite{4160945}                             & 96.30\%                                \\
LBP-TOP \cite{4160945}                             & 96.00\%                                 \\
KDT-MD \cite{4270021}            & 97.50\%                             \\
MBSIF-TOP \cite{6920055}         & 98.75\%                               \\ \midrule
Proposed PHD-VLBP, $P=3$         & 98.65\%                     \\ 
Proposed PHD-VLBP, $P=5$          & 98.50\%                     \\ 
Proposed PHD-MVLBP         & \textbf{98.90\%}                     \\ \bottomrule
\end{tabular}
}
\end{table}

\section{Conclusion}
\label{sec:majhead}
To tackle the problems of high dimensionality of VLBP and the potential information loss of LBP-TOP, we propose PHD-MVLBP for dynamic texture recognition. Firstly, PDVs are extracted and encoded into binary vectors using hash functions, where binary vectors and hash functions are jointly optimized. Then, a dictionary is derived from the binary vectors and used to encode the vectors. Multi-scale features are extracted for the neighborhood of different sizes. The proposed method could effectively extract spatial-temporal discriminant information from videos. It is evaluated on Dyntex++ and UCLA datasets and demonstrates superior performance compared with the state-of-the-art methods.

\vfill
\pagebreak
\newpage
\balance
{\small
\bibliographystyle{IEEEbib}
\bibliography{mybib}

\begin{thebibliography}{10}

\bibitem{RelevanceFeedback}
Micha Haas, Joachim Rijsdam, Bart Thomee, and Michael Lew,
\newblock ``Relevance feedback: perceptual learning and retrieval in
  bio-computing, photos, and video,''
\newblock 2004, pp. 151--156.

\bibitem{5208233}
G.~{Zhao}, M.~{Barnard}, and M.~{Pietikainen},
\newblock ``Lipreading with local spatiotemporal descriptors,''
\newblock {\em IEEE Transactions on Multimedia}, vol. 11, no. 7, pp.
  1254--1265, 2009.

\bibitem{7851001}
X.~{Li}, X.~{Hong}, A.~{Moilanen}, X.~{Huang}, T.~{Pfister}, G.~{Zhao}, and
  M.~{Pietikäinen},
\newblock ``Towards reading hidden emotions: A comparative study of spontaneous
  micro-expression spotting and recognition methods,''
\newblock {\em IEEE Transactions on Affective Computing}, vol. 9, no. 4, pp.
  563--577, 2018.

\bibitem{8600380}
X.~{Zhao}, Y.~{Lin}, L.~{Liu}, J.~{Heikkilä}, and W.~{Zheng},
\newblock ``Dynamic texture classification using unsupervised {3D} filter
  learning and local binary encoding,''
\newblock {\em IEEE Transactions on Multimedia}, vol. 21, no. 7, pp.
  1694--1708, 2019.

\bibitem{QUAN201785}
Yuhui Quan, Yuping Sun, and Yong Xu,
\newblock ``Spatiotemporal lacunarity spectrum for dynamic texture
  classification,''
\newblock {\em Computer Vision and Image Understanding}, vol. 165, pp. 85--96,
  2017.

\bibitem{Ramirez2015}
Adin Ramirez~Rivera and Oksam Chae,
\newblock ``Spatiotemporal directional number transitional graph for dynamic
  texture recognition,''
\newblock {\em IEEE Transactions on Pattern Analysis and Machine Intelligence},
  vol. 37, no. 10, pp. 2146--2152, 2015.

\bibitem{7780409}
Yuhui Quan, Chenglong Bao, and Hui Ji,
\newblock ``Equiangular kernel dictionary learning with applications to dynamic
  texture analysis,''
\newblock in {\em 2016 IEEE Conference on Computer Vision and Pattern
  Recognition (CVPR)}, 2016, pp. 308--316.

\bibitem{QI20161230}
Xianbiao Qi, Chun-Guang Li, Guoying Zhao, Xiaopeng Hong, and Matti
  Pietikäinen,
\newblock ``Dynamic texture and scene classification by transferring deep image
  features,''
\newblock {\em Neurocomputing}, vol. 171, pp. 1230--1241, 2016.

\bibitem{shervin2017dynamic}
Shervin {Rahimzadeh Arashloo}, Mehdi {Chehel Amirani}, and Ardeshir Noroozi,
\newblock ``Dynamic texture representation using a deep multi-scale
  convolutional network,''
\newblock {\em Journal of Visual Communication and Image Representation}, vol.
  43, pp. 89--97, 2017.

\bibitem{4160945}
G.~{Zhao} and M.~{Pietikainen},
\newblock ``Dynamic texture recognition using local binary patterns with an
  application to facial expressions,''
\newblock {\em IEEE Transactions on Pattern Analysis and Machine Intelligence},
  vol. 29, no. 6, pp. 915--928, 2007.

\bibitem{MBSIFTOP}
Shervin Arashloo and Josef Kittler,
\newblock ``Dynamic texture recognition using multiscale binarized statistical
  image features,''
\newblock {\em IEEE Transactions on Multimedia}, vol. 16, no. 8, pp.
  2099--2109, 2014.

\bibitem{LPQTOP}
Esa Rahtu, Janne Heikkilä, Ville Ojansivu, and Timo Ahonen,
\newblock ``Local phase quantization for blur-insensitive image analysis,''
\newblock {\em Image and Vision Computing}, vol. 30, no. 8, pp. 501–512,
  2012.

\bibitem{ren2013dynamic}
Jianfeng Ren, Xudong Jiang, and Junsong Yuan,
\newblock ``Dynamic texture recognition using enhanced {LBP} features,''
\newblock in {\em 2013 IEEE International Conference on Acoustics, Speech and
  Signal Processing}, 2013, pp. 2400--2404.

\bibitem{ren2015chi}
Jianfeng Ren, Xudong Jiang, and Junsong Yuan,
\newblock ``A {Chi}-squared-transformed subspace of {LBP} histogram for visual
  recognition,''
\newblock {\em IEEE Transactions on Image Processing}, vol. 24, no. 6, pp.
  1893--1904, 2015.

\bibitem{ren2015learning}
Jianfeng Ren, Xudong Jiang, and Junsong Yuan,
\newblock ``Learning {LBP} structure by maximizing the conditional mutual
  information,''
\newblock {\em Pattern Recognition}, vol. 48, no. 10, pp. 3180--3190, 2015.

\bibitem{Hong2018}
Sungeun Hong, Jongbin Ryu, and Hyun~S. Yang,
\newblock ``Not all frames are equal: aggregating salient features for dynamic
  texture classification,''
\newblock {\em Multidimensional Systems and Signal Processing}, vol. 29, no. 1,
  pp. 279--298, 2018.

\bibitem{ren2013noise}
Jianfeng Ren, Xudong Jiang, and Junsong Yuan,
\newblock ``Noise-resistant local binary pattern with an embedded
  error-correction mechanism,''
\newblock {\em IEEE Transactions on Image Processing}, vol. 22, no. 10, pp.
  4049--4060, 2013.

\bibitem{1017623}
T.~{Ojala}, M.~{Pietikainen}, and T.~{Maenpaa},
\newblock ``Multiresolution gray-scale and rotation invariant texture
  classification with local binary patterns,''
\newblock {\em IEEE Transactions on Pattern Analysis and Machine Intelligence},
  vol. 24, no. 7, pp. 971--987, 2002.

\bibitem{ren2015quantized}
Jianfeng Ren, Xudong Jiang, and Junsong Yuan,
\newblock ``Quantized fuzzy {LBP} for face recognition,''
\newblock in {\em 2015 IEEE International Conference on Acoustics, Speech and
  Signal Processing (ICASSP)}, 2015, pp. 1503--1507.

\bibitem{ren2017sound}
Jianfeng Ren, Xudong Jiang, Junsong Yuan, and Nadia Magnenat-Thalmann,
\newblock ``Sound-event classification using robust texture features for robot
  hearing,''
\newblock {\em IEEE Transactions on Multimedia}, vol. 19, no. 3, pp. 447--458,
  2017.

\bibitem{7936534}
Y.~{Duan}, J.~{Lu}, J.~{Feng}, and J.~{Zhou},
\newblock ``Context-aware local binary feature learning for face recognition,''
\newblock {\em IEEE Transactions on Pattern Analysis and Machine Intelligence},
  vol. 40, no. 5, pp. 1139--1153, 2018.

\bibitem{2survey}
di~Huang, Caifeng Shan, Mohsen Ardabilian, and Liming Chen,
\newblock ``Local binary patterns and its application to facial image analysis:
  A survey,''
\newblock {\em IEEE Transactions on Systems, Man, and Cybernetics, Part C},
  vol. 41, no. 6, pp. 765--781, 2011.

\bibitem{ghanem2010maximum}
Bernard Ghanem and Narendra Ahuja,
\newblock ``Maximum margin distance learning for dynamic texture recognition,''
\newblock in {\em Computer Vision -- ECCV 2010}, Kostas Daniilidis, Petros
  Maragos, and Nikos Paragios, Eds., Berlin, Heidelberg, 2010, pp. 223--236,
  Springer Berlin Heidelberg.

\bibitem{dyntex}
Renaud Péteri, Sándor Fazekas, and Mark~J. Huiskes,
\newblock ``Dyntex: A comprehensive database of dynamic textures,''
\newblock {\em Pattern Recognition Letters}, vol. 31, no. 12, pp. 1627--1632,
  2010.

\bibitem{6920055}
S.~R. {Arashloo} and J.~{Kittler},
\newblock ``Dynamic texture recognition using multiscale binarized statistical
  image features,''
\newblock {\em IEEE Transactions on Multimedia}, vol. 16, no. 8, pp.
  2099--2109, 2014.

\bibitem{8030131}
X.~{Zhao}, Y.~{Lin}, and J.~{Heikkilä},
\newblock ``Dynamic texture recognition using volume local binary count
  patterns with an application to {2D} face spoofing detection,''
\newblock {\em IEEE Transactions on Multimedia}, vol. 20, no. 3, pp. 552--566,
  2018.

\bibitem{10.1109}
Yong Xu, Yuhui Quan, Haibin Ling, and Hui Ji,
\newblock ``Dynamic texture classification using dynamic fractal analysis,''
\newblock in {\em Proceedings of the IEEE International Conference on Computer
  Vision}, 2011, pp. 1219--1226.

\bibitem{4270021}
A.~B. {Chan} and N.~{Vasconcelos},
\newblock ``Classifying video with kernel dynamic textures,''
\newblock in {\em 2007 IEEE Conference on Computer Vision and Pattern
  Recognition}, 2007, pp. 1--6.

\end{thebibliography}
}

\end{document}